\pdfoutput=1

\documentclass[11pt]{article}

\usepackage[]{ACL2023}

\usepackage{times}
\usepackage{latexsym}
\usepackage{multirow}
\usepackage{enumitem}
\usepackage{algorithm}
\usepackage{algpseudocode}
\usepackage{booktabs}

\usepackage{graphicx}
\usepackage{amssymb}
\usepackage{makecell}
\usepackage{amsmath, bm}
\usepackage{longtable}
\usepackage{caption}
\usepackage{float}
\usepackage{xcolor}
\usepackage{microtype}

\usepackage[T1]{fontenc}

\usepackage[utf8]{inputenc}

\usepackage{microtype}

\usepackage{inconsolata}

\title{Learning In-context Learning for Named Entity Recognition}

\author{
  Jiawei Chen${}^{1,4,}$\thanks{~ This work was partially done when Jiawei Chen interned at Baidu.},
  Yaojie Lu${}^{1,}$\thanks{~ Corresponding authors.},
  Hongyu Lin${}^{1}$,
  Jie Lou${}^{3}$,
  Wei Jia${}^{3}$,
  Dai Dai${}^{3}$,
  \\
  {\bf Hua Wu${}^{3}$},
  {\bf Boxi Cao${}^{1,4}$},
  {\bf Xianpei Han${}^{1,2,}$\footnotemark[2]},
  {\bf Le Sun${}^{1,2}$}
  \\
  ${}^{1}$Chinese Information Processing Laboratory ~
  ${}^{2}$State Key Laboratory of Computer Science \\
  Institute of Software, Chinese Academy of Sciences, Beijing, China\\
  ${}^{3}$Baidu Inc., Beijing, China \\
  ${}^{4}$University of Chinese Academy of Sciences, Beijing, China \\
  {\tt \{jiawei2020,yaojie,hongyu,boxi2020,xianpei,sunle\}@iscas.ac.cn} \\
  {\tt \{loujie,jiawei07,daidai,wu\_hua\}@baidu.com} \\
}

\begin{document}
\maketitle
\begin{abstract}
Named entity recognition in real-world applications suffers from the diversity of entity types, the emergence of new entity types, and the lack of high-quality annotations. To address the above problems, this paper proposes an in-context learning-based NER approach, which can effectively inject in-context NER ability into PLMs and recognize entities of novel types on-the-fly using only a few demonstrative instances. Specifically, we model PLMs as a meta-function $\mathcal{ \lambda_  {\text{instruction, demonstrations, text}}. M}$\footnote{This paper represents functions using lambda-calculus~\citep{barendregt1992lambda}, and each function is represented as $\mathcal{\lambda}_{x,y,z}. M$, where $x,y,z$ are variables and $M$ is function definition/abstraction. The function can apply to arguments such as $\mathcal{ (\lambda}_{x,y,z}. M)(x=A,y=B, z=C)$ (fully applied) or $\mathcal{ (\lambda}_{x,y,z}. M)(x=A,y=B)$ (partially applied).}, and a new entity extractor can be implicitly constructed by applying new instruction and demonstrations to PLMs, i.e., $\mathcal{ (\lambda . M) }$(instruction, demonstrations) $\to$ $\mathcal{F}$ where $\mathcal{F}$ will be a new entity extractor, i.e., $\mathcal{F}$: text $\to$ entities. To inject the above in-context NER ability into PLMs, we propose a meta-function pre-training algorithm, which pre-trains PLMs by comparing the (instruction, demonstration)-initialized extractor with a surrogate golden extractor. Experimental results on 4 few-shot NER datasets show that our method can effectively inject in-context NER ability into PLMs and significantly outperforms the PLMs+fine-tuning counterparts.

\end{abstract}

\section{Introduction}
Named entity recognition (NER) aims to detect and classify named entities in text, such as \textit{People}, \textit{Disease}, and \textit{Movie}. Traditional NER methods~\citep{lample-etal-2016-neural,li-etal-2020-unified,yan-etal-2021-unified-generative} have achieved remarkable success when entity types are pre-defined and massive high-quality annotations are provided. Unfortunately, real-world NER still suffers from the diversity of entity types (e.g., the extraction of \textit{Movie} is very different to \textit{Disease}), the emergence of new entity types (e.g.,  \textit{Virus} of Cov-19 ), and the lack of high-quality annotations.

\begin{figure}[t!]
\centering 
\includegraphics[width=0.99\columnwidth]{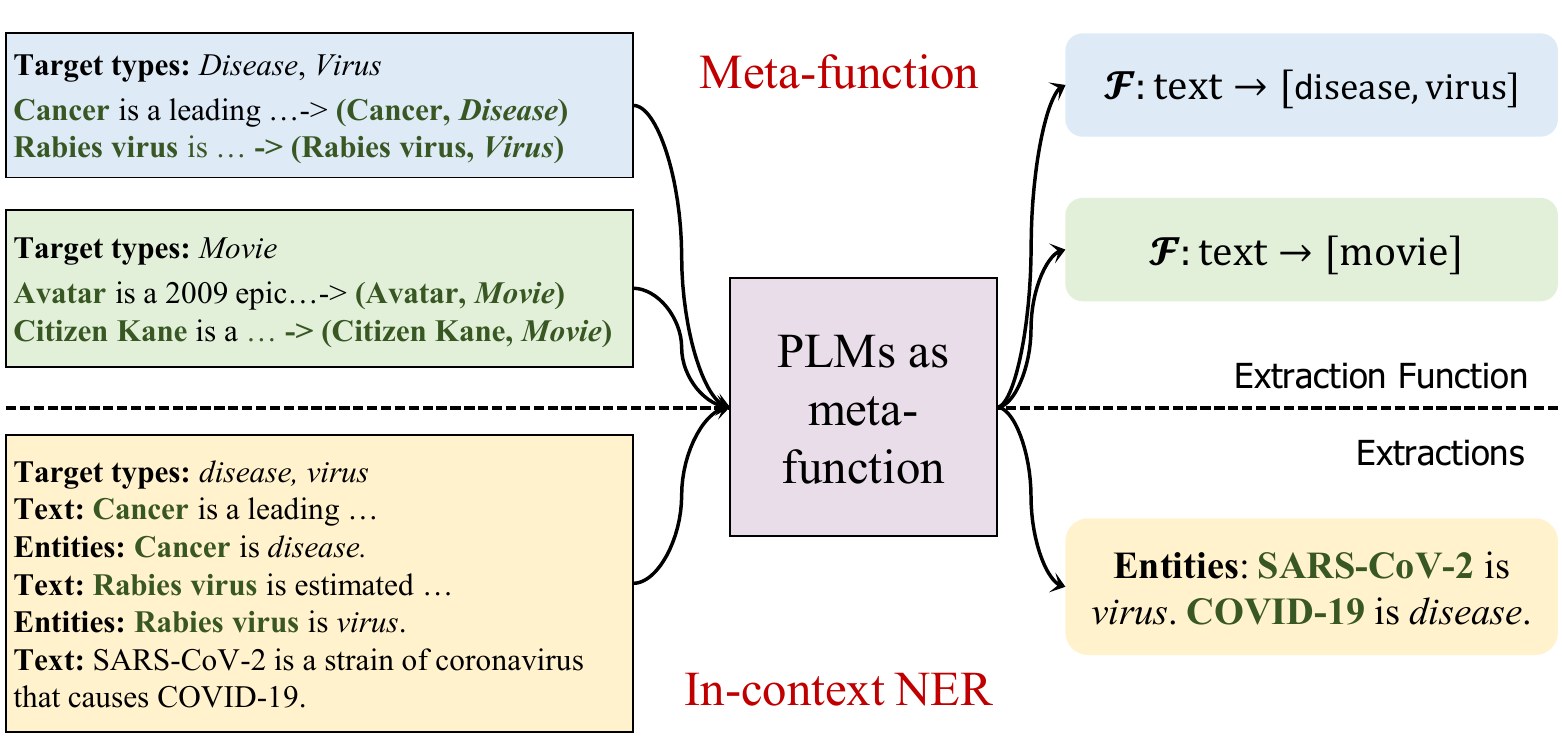}
\caption{Illustration of in-context NER, which uses instruction, demonstrations, and text as input to identify entities. The in-context learning model can be regarded as a meta-function that takes instruction and demonstrations as input and produces an entity extractor capable of identifying the desired entities~\citep{akyurek2022learning}.}
\label{Fig:exam}
\end{figure}

To address these problems, recent studies often employ few-shot learning techniques, including fine-tuning-based and metric-based methods.
Fine-tuning-based methods extract entities of new types by adjusting model weights using new instances ~\citep{ma-etal-2022-label,chen-etal-2022-shot,das-etal-2022-container}.
The main drawbacks of these methods are that re-training is often expensive (especially for large-scale models) and new entity types cannot be addressed on-the-fly.
Metric-based methods are free from updating parameters and identifying entities by learning to compare query instances with support instances (or prototypes)~\citep{yang-katiyar-2020-simple,tong-etal-2021-learning}. These methods are limited to the matching architectures and are sensitive to domain shift since they do not fully explore the information of target domain~\citep{ma-etal-2022-decomposed}.

In this paper, we propose an in-context learning-based NER approach, which can effectively address the above problems by injecting in-context NER ability into PLMs and then recognizing entities of new types on-the-fly using only a few demonstrative instances. 
Specifically, we model PLMs as a meta-function ~\citep{akyurek2022learning} for NER $\mathcal{ \lambda_ {\text{instruction, demonstrations, text}}. M}$, and a new entity extractor can be implicitly constructed by applying new instruction and demonstrations to PLMs, i.e., $\mathcal{ (\lambda . M) }$(instructions, demonstrations) $\to$ $\mathcal{F}$ where $\mathcal{F}$ will be a new entity extractor $\mathcal{F}$: text $\to$ entities. For example, in Figure~\ref{Fig:exam}, our method can construct entity extractors of new \emph{Disease} and \emph{Virus} types on-the-fly by applying PLMs using demonstrations such as ``Text: Cancer is a leading cause of death worldwide. Entities: Cancer is disease''. 
Furthermore, we propose a meta-function pre-training algorithm to inject the above in-context NER ability into PLMs. The algorithm pre-trains PLMs by comparing the implicitly (instruction, demonstration)-constructed extractor with an explicitly fine-tuned surrogate golden extractor. The comparison ensures that the meta-function $\mathcal{ (\lambda.M) }$ will generate an entity extractor $\mathcal{F}$ from instructions and demonstrations as accurately as possible.

The proposed method can seamlessly leverage the powerful language understanding and generation capabilities of large-scale PLMs~\citep{NEURIPS2020_1457c0d6}, effectively address diverse and new entity types through in-context learning, and only requires a couple of demonstrations for each entity type.
Compared to fine-tuning methods, our method does not require expensive retraining, and new entity types can be extracted on-the-fly, with no need for model weight adjusting. 
Compared with metric-based methods, our method can dynamically utilize the information entailed in instruction and demonstrations rather than be limited to the fixed metric space. 
 
To verify the effectiveness of our method, we further pre-train PLMs using a large-scale distantly annotated NER dataset from Wikipedia and Wikidata. Experimental results on 4 few-shot NER benchmarks show that our method can effectively inject in-context NER ability into PLMs and significantly outperforms the PLMs+fine-tuning counterparts\footnote{Our source codes are openly available at \url{https://github.com/chen700564/metaner-icl}}.

In general, this paper's main contributions are:
\begin{itemize} 
    \item We propose an in-context NER method that can effectively extract entities of novel types on-the-fly using only a few demonstrative  instances.
    
    \item We design a meta-function pre-training algorithm, which models PLMs as a meta-function and injects in-context NER ability into PLMs by comparing the (instruction, demonstration)-constructed extractor with a surrogate golden extractor. 
    
    \item How to inject in-context ability into small models is an important research direction of NLP in the big model era. Our work can benefit new directions for future works.
\end{itemize}
\section{Related work}
\paragraph{Few-shot NER} Few-shot learning is a promising technique for low-resource NER. Currently, there are two main categories of FS-NER methods: fine-tuning-based methods and metric-based methods. Fine-tuning-based FS-NER methods re-train NER models using new instances. Metric-based methods identify entities by pre-training to compare query instances with support instances~\citep{snell2017prototypical, fritzler2019few,yang-katiyar-2020-simple,tong-etal-2021-learning,wang-etal-2022-enhanced,ji-etal-2022-shot} using given NER datasets. FS-NER is a challenging task, and several improvements have been proposed to enhance its performance. These include leveraging label information ~\citep{DBLP:conf/acl/HouCLZLLL20,wang-etal-2021-learning-language-description,lu-etal-2022-unified,ma-etal-2022-label,chen-etal-2022-shot,yang-etal-2022-see}, designing new paradigms such as decomposition methods~\citep{ji-etal-2022-shot,ma-etal-2022-decomposed,yang-etal-2022-see}, prompt-based methods~\citep{cui-etal-2021-template,liu2022qaner,ma-etal-2022-template}, and demonstration-based methods~\citep{lee-etal-2022-good,zhang2022robustness}; , and proposing new learning strategies like meta-learning~\citep{li2020few,li2020metaner,de-lichy-etal-2021-meta,ma-etal-2022-decomposed}, contrastive learning~\citep{das-etal-2022-container}, and self-training~\citep{huang-etal-2021-shot,wang2021meta}. In this paper, we address FS-NER via in-context learning~\citep{gutierrez2022thinking}, which empowers PLMs with in-context learning ability and entities of new 
entity types can be extracted on-the-fly.

\paragraph{In-context learning} The in-context learning ability has been observed in large-scale PLMs such as GPT-3~\citep{NEURIPS2020_1457c0d6}, and has been widely applied in different tasks such as ``chain of thought'' reasoning ~\citep{wei2022chain}. Recent studies aim to enhance in-context learning by selecting valuable demonstrations~\citep{liu2021makes,rubin-etal-2022-learning}, optimizing the order of demonstrations~\citep{lu-etal-2022-fantastically}, and calibrating output distributions~\citep{zhao2021calibrate}. Some studies try to replicate in-context learning in smaller models~\citep{min-etal-2022-metaicl,chen-etal-2022-improving}. Additionally, some researchers attempt to replicate in-context learning using smaller models~\citep{min2022rethinking, chan2022data}. Furthermore, there are efforts to understand the underlying mechanisms~\citep{akyurek2022learning} of in-context learning which suggest that it can be compared to a meta-function and facilitate implicit fine-tuning~\citep{dai2022can,von2022transformers}. This paper is inspired by previous studies and considers in-context named entity recognition (NER) as a meta-function. To enhance the ability of pre-trained language models (PLMs) to perform in-context NER, we propose an effective pre-training algorithm. Unlike MetaICL~\citep{min-etal-2022-metaicl}, which only transforms multi-task learning into the form of in-context learning for pre-training, our approach also includes meta-function pre-training~(Section~\ref{sec:ef}) based on the underlying mechanisms of in-context learning.

\section{In-context Named Entity Recognition}\label{sec:form}

\begin{figure}[t!]
\centering 
\includegraphics[width=0.42\textwidth]{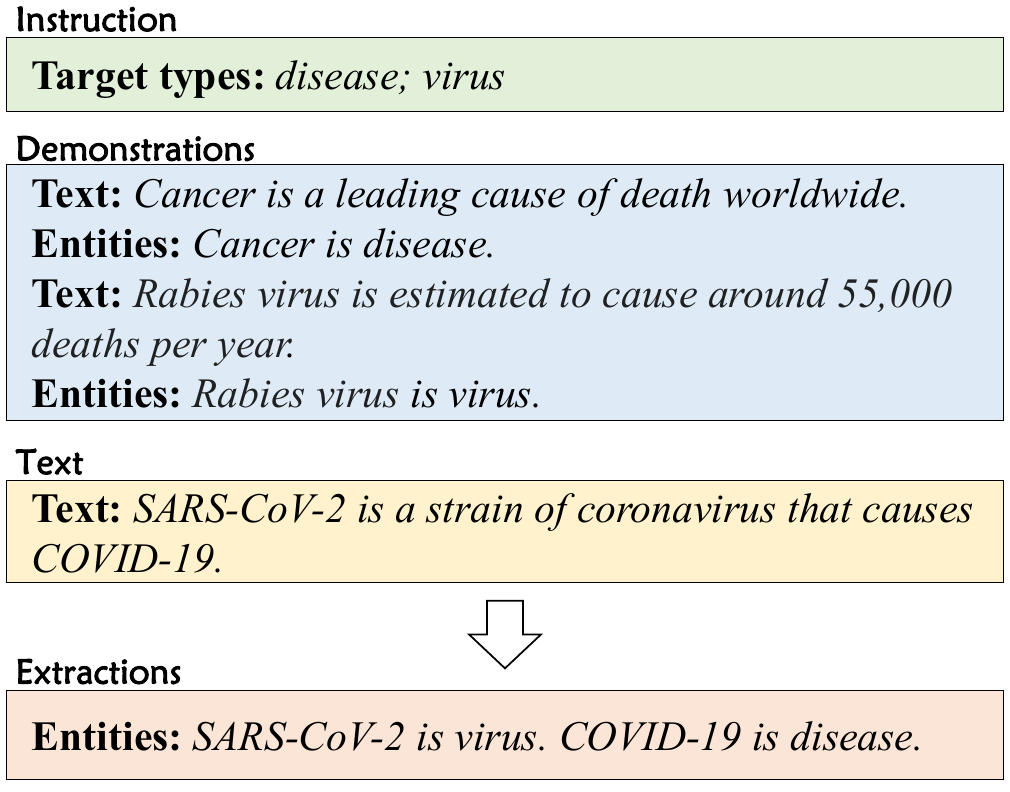}
\caption{The formats of input and output of in-context few-shot NER. The input is formed by instruction, demonstrations, and text.}
\label{Fig:io}
\end{figure}

This section describes how to recognize entities through in-context NER. In in-context learning, the model will read the information of target entity types from both instruction and demonstrations, and then extract entities of target types within the text. In this way, new entity types can be extracted on-the-fly, without the need for model retraining.

Concretely, this paper formulates in-context NER as a sequence-to-sequence generation process. The input  $X = [I; D; T]$ includes instruction $I$, demonstrations $D$, and text $T$ while the output is a list of extracted entities $Y=[e_1, ... , e_n]$. Figure~\ref{Fig:io} shows an example, where an in-context NER model will identify that the target entity types are {\emph{Disease} and \emph{Virus}}, distill the knowledge about these types from demonstrations(e.g., the context patterns of a disease), and finally recognize "SARS-CoV-2" as virus and ``COVID-19'' as disease using the above knowledge. The details are described as follows.

\paragraph{Instruction} The instruction is a sequence of target entity types, guiding the model to extract what entity types ~\citep{min-etal-2022-metaicl}. The instruction for target entity types $\{l_1,\dots,l_n\}$ is $I=$``Target types: $l_1; \dots; l_n$''. For example, in Figure~\ref{Fig:io} the instruction is ``Target types: disease; virus''.

\paragraph{Demonstrations}
Demonstrations provide the intra-class knowledge of target entity types (e.g., entity semantics and context patterns) and illustrate the form of outputs. As shown in Figure~\ref{Fig:io}, the demonstrations contain the illustrative instances for different target types, and each instance is ``Text: \{text\} Entities: \{extractions\}'', where \{extractions\} are entities presented in the \{text\}.

\paragraph{Extractions}
The output of the extraction process is a list of entities, denoted as $Y=[e_1, \dots, e_n]$ where $e_i$ is $i$-th extracted entities. Each extraction $e$ is represented as ``\textsc{Entity} is \textit{type}''. For instance, in Figure~\ref{Fig:io}, the extraction ``COVID-19 is disease.'' indicates that ``COVID-19'' is an entity mention with the type ``Disease''. This natural language-like representation allows us to better utilize the text generation capabilities of pre-trained language models. During inference, we locate all mentions in the text and further output their locations.

\paragraph{Architecture} 
Given the above task formulation, we employ an encoder-decoder network like T5~\citep{raffel2020exploring}, where the encoder encodes <instruction, demonstrations, text> and the decoder generates all extractions as a tokenized text sequence $Y=[y_1, \dots, y_n]$.

The success of in-context NER depends on two critical abilities: the in-context learning ability and the extraction ability. For in-context learning, the models should be able to implicitly construct accurate extractors of new entity types by following the instruction and capturing the knowledge in demonstrations. In this way, we can see a PLM as a meta-function, i.e., a function of extractors whose input is (instruction, demonstrations) and whose output is an entity extractor. For extraction, the models should be able to locate specific spans and categorize them into target entity types. The following section demonstrates how to inject such an in-context learning ability into PLMs and construct an effective in-context NER model.

\section{Meta-Function Pre-training for In-Context NER}
In this section, we will explain how to incorporate in-context named entity recognition (NER) capabilities into pre-trained language models (PLMs). Although large-scale PLMs like GPT-3 have demonstrated the ability to learn in-context, this capability is not always controllable or predictable. Additionally, unlike classification and question-answering tasks that align with the pre-training objective of language models (i.e., producing natural text output), NER requires more complex span extraction and type specification.
As a result, \citet{gutierrez2022thinking} show that LMs aren’t well-suited for in-context NER tasks. In this paper, we propose meta-function pre-training, an algorithm that can inject in-context NER ability into PLMs in a controllable and predictable way.

Specifically, we model PLMs as a meta-function ~\citep{akyurek2022learning} for NER $\mathcal{ \lambda_  {\text{instruction, demonstrations, text}}. M}$, and a new entity extractor can be implicitly constructed by applying new instructions and demonstrations to PLMs, i.e., $\mathcal{ (\lambda . M) }$(instructions, demonstractions) $\to$ $\mathcal{F}$ where $\mathcal{F}$ will be a new entity extractor $\mathcal{F}$:text $\to$ entities. Based on the meta-function formulation, we further pre-train PLMs for in-context NER abilities by:
\begin{itemize}[nosep]
    \item optimizing PLMs via a meta-function loss, so that the implicitly (instruction, demonstration)-constructed extractor $\mathcal{F}$ will be as close as an explicitly fine-tuned surrogate golden extractor;
    \item optimizing PLMs via an extraction loss, so that the in-context NER can effectively locate and categorize entities in a text.
\end{itemize} 
The details are described in the following.

\subsection{Pre-training Settings}

\paragraph{Pre-training Corpus Construction}
To continually pre-train PLMs for in-context NER, we first collect an in-context pre-training  NER corpus $\mathcal{D}_{\text{in-context}} = \{x_1, x_2, ..., x_n \}$, where each $x$ is an in-context NER task represented as a tuple   = (instruction, demonstrations, text, entities). 

Specifically, to sample in-context NER task $x$, we use traditional NER corpus $\mathcal{D}_{\text{NER}}$ where each NER instance is a (text, entities) pair as follows:
\begin{enumerate}[nosep,leftmargin=*]
    \item \textbf{In-context Task Sampling}:
    To construct an in-context NER task x = (instruction, demonstrations, text, entities): (1) we first sample $N$ target entity types from $\mathcal{D}_{\text{NER}}$ to form instruction and sample $K$ instances for each type to form demonstrations; (2) then we sample the text and the entities of x by either randomly sample an instance from $N$ target entity types, or randomly sample from instances of other entity types, i.e., their extractions are NIL. We sample NIL instances because in real-world applications many instances will not contain target entities, and NIL instances are sampled with a predefined proportion $\gamma$.
    
    \item \textbf{Type Anonymization}: To ensure the models rely on in-context demonstrations for entity knowledge and avoid overfitting to entity type names, we anonymize entity types by randomly substituting them with a set of type indicators \{<type1>, $\ldots$, <type99>\}, rather than directly using the original type names such as ~\emph{Disease} and ~\emph{Virus}. We found this anonymization strategy can significantly improve the in-context learning ability of PLMs. 
    Specifically, we randomly substitute each entity type name with pre-defined 99 type indicators \{<type1>, $\ldots$, <type99>\}, and the substitute probability for each name is 80\%.
\end{enumerate}

\paragraph{Pre-training Loss}
Based on the in-context pre-training corpus $\mathcal{D}_{\text{in-context}}$, we pre-train our in-context NER model by optimizing the loss:
\begin{equation}
\mathcal{L}=\alpha \mathcal{L}_{\text{meta-function}} + \mathcal{L}_{\text{extraction}}
\end{equation}
where $\mathcal{L}_{\text{meta-function}}$ is the meta-function loss which ensures PLMs can implicitly generate accurate entity extractors (Section~\ref{sec:mf}),  $\mathcal{L}_{\text{extraction}}$ is the extraction loss which ensures PLMs have good extraction ability (Section~\ref{sec:ef}), $\alpha$ is the coefficient of meta-function loss.

\subsection{Meta-function Pre-training}\label{sec:mf}

\begin{figure*}[t!]
\centering 
\includegraphics[width=0.95\textwidth]{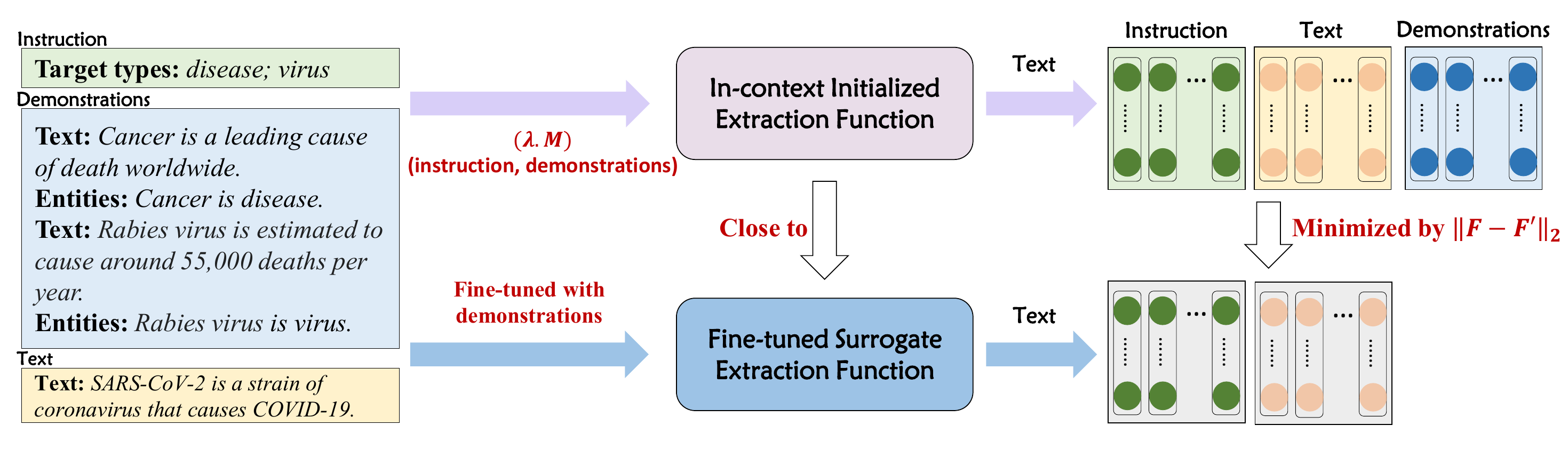}
\caption{Overview of our meta-function pre-training. Our goal is to ensure that the extractor $\mathcal{F}$(instruction,demonstrations) closely resembles the golden extraction function. To obtain the golden extraction function, we use a surrogate strategy and the surrogate extraction function is the fine-tuned encoder using demonstrations.}
\label{Fig:model}
\end{figure*}

As mentioned above, a good in-context NER model should be able to implicitly construct an accurate entity extractor by partially applying PLMs with instruction $I$ and demonstrations $D$: 
\begin{equation}
    (\lambda.M)(I, D) = \mathcal{F}
\end{equation}
For example, given the instruction and demonstrations in Figure 2, we want PLMs to implicitly build an accurate extractor for \emph{Disease} and \emph{Virus}. Therefore if we know the golden extraction function $\mathcal{F}^*$ for target entity types, we can optimize PLMs for in-context NER ability by minimizing the distance ||$\mathcal{F}^* - \mathcal{F}$||.

Unfortunately, the golden extraction function $\mathcal{F}^*$ is unknown. In this paper, we approximate $\mathcal{F}^*$ using a surrogate extractor which is the fine-tuned counterpart using demonstrations $D$. That is, for each in-context pre-training task $x$, we first recover all NER (text, entities) instances from $x$ as $x'$, then we fine-tune the model and use the fine-tuned encoder $\mathcal{F}'$ as the surrogate of $\mathcal{F}^*$.
The overall meta-function pre-training is shown in Figure~\ref{Fig:model}.

Formally, given instruction $I$, demonstration $D$, and text $T$, we first feed them into the encoder and obtain the feature of $I$ and $T$,
\begin{equation}\small
\mathbf{l}_1,...,\mathbf{l}_n,\mathbf{d}_1,...,\mathbf{d}_m,\mathbf{t}_1,...,\mathbf{t}_k= Encoder(I; D; T)
\end{equation}
Then we obtain the feature of the implicitly generated function $\mathcal{F}$ using the features of instruction $I$ and text $T$, and ignore the features of $D$: $\mathbf{F} = [\mathbf{l}_1,...,\mathbf{l}_n, \mathbf{t}_1,...,\mathbf{t}_k]$.
In Figure~\ref{Fig:model}, the feature $\mathbf{F}$ can be seen as the output of \emph{Disease} and \emph{Virus} extractor $\mathcal{F}$.

To obtain the feature of the fine-tuned counterpart, we perform a one-step gradient descent\footnote{We use “one-step gradient” because it strikes a balance between cost and effectiveness. While we believe that more steps may result in better performance, it would require additional time which we plan to explore in future work.} on the encoder using the instances in the demonstration $D$ and get the surrogate encoder, which can be seen as an approximation of golden $\mathcal{F}^*$. Note that this fine-tuning operation is performed after the model has been copied, so there is no impact on the parameters of the original model. In the example in \figurename~\ref{Fig:model}, $Encoder'$ is a \emph{Disease} and \emph{Virus} extractor.
After performing one-step updating, we feed instruction and text $[I; T]$ into the surrogate encoder to get their features:
\begin{equation}
\mathbf{F}' = Encoder'(I; T)
\end{equation}
where $\mathbf{F}'=\{\mathbf{l_1'},\ldots,\mathbf{l_n'},\mathbf{t_1'},\ldots,\mathbf{t_k'}\}$ is features of instruction $I$ and text $T$. In the example in Figure~\ref{Fig:model}, the feature $\mathbf{F'}$ can be seen as the estimated output of golden extractor $\mathcal{F}^*$ for \emph{Virus} and \emph{Disease} entity types. 

Then, we pre-train our in-context NER model to be a good meta-function by making the output of $F$ and $F^*$ consistent, i.e., minimizing the distance between $\mathbf{F}$ and $\mathbf{F'}$. The meta-function loss is:
\begin{equation}
\mathcal{L}_{\text{meta-function}}= \frac{1}{n+k}\sum_{i=1}^{n+k}d(\mathbf{F}_{i},\mathbf{F'}_{i})
\end{equation}
where $d(\cdot)$ is euclidean distance. Note that when calculating the gradient of $\mathcal{L}_{\text{meta-function}}$, $\mathbf{F'}$ is seen as constant. To this end, the meta-function gradient can be estimated as:
\begin{equation}
\nabla \theta_{\text{encoder}}= \frac{\partial \mathcal{L}_{\text{meta-function}}}{\partial X}
\end{equation}
where $\theta_{\text{encoder}}$ is the parameters of the encoder and $X=[I;D;T]$ is the input. The estimated gradient will be used to update the parameters of the encoder.

In this way, the in-context NER models will be trained to be a good meta-function ~\citep{akyurek2022learning}, which can also be seen as an ability for implicit fine-tuning~\citep{dai2022can,von2022transformers}.

\subsection{Extraction Function Pre-training}\label{sec:ef}

Besides the in-context learning ability, we also pre-train PLMs to be good extractors via extraction loss. 
Given instruction $I$, demonstrations $D$, and text $T$, the sequence-to-sequence entity extractor directly models the generation probability token by token in an auto-regressive way.
Formally, we optimize the model parameters $\theta$ by minimizing the negative likelihood of in-context instances:
\begin{equation}
\mathcal{L}_{\text{extraction}} = - \log \prod_{i=1}^{|Y|} P(y_{i}|y_{<i},X,\theta)
\end{equation}
And the extraction gradient is computed as:
\begin{equation}
\nabla \theta= \frac{\partial \mathcal{L}_{\text{extraction}}}{\partial X}
\end{equation}

\begin{table*}[th!]
\centering
\resizebox{0.9\textwidth}{!}{
\begin{tabular}{@{}lcccccccccc@{}}
\toprule
\multicolumn{1}{c|}{\multirow{2}{*}{Models}}                     & \multicolumn{1}{c|}{\multirow{2}{*}{\#Param}} & \multicolumn{2}{c|}{CoNLL03}                         & \multicolumn{2}{c|}{WNUT17}                          & \multicolumn{2}{c|}{NCBI-disease}                    & \multicolumn{2}{c|}{SEC-filings}                     & \multirow{2}{*}{AVE} \\ \cmidrule(lr){3-10}
\multicolumn{1}{c|}{}                                            & \multicolumn{1}{c|}{}                         & 1-shot         & \multicolumn{1}{c|}{5-shot}         & 1-shot         & \multicolumn{1}{c|}{5-shot}         & 1-shot         & \multicolumn{1}{c|}{5-shot}         & 1-shot         & \multicolumn{1}{c|}{5-shot}         &                      \\ \midrule
\multicolumn{11}{c}{\textbf{Pre-trained Language Models}}                                                                                                                                                                                                                                                                                                           \\ \midrule
\multicolumn{1}{l|}{T5v1.1-large}     & \multicolumn{1}{c|}{770M}                     & 38.61          & \multicolumn{1}{c|}{44.90}          & 25.52          & \multicolumn{1}{c|}{26.32}          & 26.02          & \multicolumn{1}{c|}{37.63}          & 41.89          & \multicolumn{1}{c|}{53.44}          & 36.79                \\
\multicolumn{1}{l|}{GPT2-xl}         & \multicolumn{1}{c|}{1.5B}                     & 33.69          & \multicolumn{1}{c|}{39.55}          & 22.63          & \multicolumn{1}{c|}{24.86}          & 25.54          & \multicolumn{1}{c|}{33.25}          & 42.83          & \multicolumn{1}{c|}{\textbf{57.05}} & 34.93                \\
\multicolumn{1}{l|}{T5-xl}           & \multicolumn{1}{c|}{3B}                       & 38.99          & \multicolumn{1}{c|}{45.74}          & 26.39          & \multicolumn{1}{c|}{26.31}          & 23.10          & \multicolumn{1}{c|}{36.78}          & 30.58          & \multicolumn{1}{c|}{42.22}          & 33.76                \\
\multicolumn{1}{l|}{GPT-J-6B}                      & \multicolumn{1}{c|}{6B}                       & 46.14          & \multicolumn{1}{c|}{50.10}          & 31.41          & \multicolumn{1}{c|}{30.93}          & 35.82          & \multicolumn{1}{c|}{40.98}          & 40.12          & \multicolumn{1}{c|}{39.61}          & 39.39                \\
\multicolumn{1}{l|}{T5-xxl}          & \multicolumn{1}{c|}{11B}                      & 40.97          & \multicolumn{1}{c|}{46.14}          & 24.76          & \multicolumn{1}{c|}{25.27}          & 12.19          & \multicolumn{1}{c|}{26.34}          & 32.65          & \multicolumn{1}{c|}{42.44}          & 31.35                \\
\multicolumn{1}{l|}{OPT-13B}                & \multicolumn{1}{c|}{13B}                      & 46.65          & \multicolumn{1}{c|}{51.71}          & 27.74          & \multicolumn{1}{c|}{28.36}          & 23.73          & \multicolumn{1}{c|}{34.00}          & 41.60          & \multicolumn{1}{c|}{43.10}          & 37.11                \\
\multicolumn{1}{l|}{GPT-Neox-20B}           & \multicolumn{1}{c|}{20B}                      & 52.68          & \multicolumn{1}{c|}{58.12}          & \textbf{36.29} & \multicolumn{1}{c|}{35.68}          & 35.42          & \multicolumn{1}{c|}{42.85}          & 45.07          & \multicolumn{1}{c|}{45.17}          & 43.91                \\
\multicolumn{1}{l|}{OPT-30B}                & \multicolumn{1}{c|}{30B}                      & 42.86          & \multicolumn{1}{c|}{44.77}          & 25.85          & \multicolumn{1}{c|}{27.44}          & 22.31          & \multicolumn{1}{c|}{32.76}          & 40.83          & \multicolumn{1}{c|}{46.52}          & 35.42                \\
\multicolumn{1}{l|}{OPT-66B}                & \multicolumn{1}{c|}{66B}                      & 43.83          & \multicolumn{1}{c|}{53.89}          & 30.77          & \multicolumn{1}{c|}{32.00}          & 25.87          & \multicolumn{1}{c|}{34.58}          & 39.15          & \multicolumn{1}{c|}{47.01}          & 38.39                \\ \midrule
\multicolumn{11}{c}{\textbf{Pre-trained NER Models}}                                                                                                                                                                                                                                                                                                                \\ 
\midrule
\multicolumn{1}{l|}{ProtoNet}      & \multicolumn{1}{c|}{345M}                     & 30.04          & \multicolumn{1}{c|}{60.26}          & 9.74           & \multicolumn{1}{c|}{23.03}          & 24.73          & \multicolumn{1}{c|}{42.32}          & 16.79          & \multicolumn{1}{c|}{23.67}          & 28.82                \\
\multicolumn{1}{l|}{NNShot}     & \multicolumn{1}{c|}{345M}                     & 41.92          & \multicolumn{1}{c|}{58.39}          & 15.76          & \multicolumn{1}{c|}{21.78}          & 31.59          & \multicolumn{1}{c|}{33.14}          & 30.19          & \multicolumn{1}{c|}{37.86}          & 33.83                \\
\multicolumn{1}{l|}{StructShot} & \multicolumn{1}{c|}{345M}                     & 42.34          & \multicolumn{1}{c|}{58.44}          & 15.78          & \multicolumn{1}{c|}{22.05}          & 19.87          & \multicolumn{1}{c|}{31.48}          & 30.40          & \multicolumn{1}{c|}{38.44}          & 32.35                \\
\multicolumn{1}{l|}{CONTAINER}   & \multicolumn{1}{c|}{345M}                     & 45.43          & \multicolumn{1}{c|}{61.69}          & 15.64          & \multicolumn{1}{c|}{20.37}          & 23.24          & \multicolumn{1}{c|}{27.02}          & 34.07          & \multicolumn{1}{c|}{40.44}          & 33.49                \\  \midrule

\multicolumn{1}{l|}{MetaNER-base}                                 & \multicolumn{1}{c|}{220M}                     & 53.94          & \multicolumn{1}{c|}{62.59}          & 25.55          & \multicolumn{1}{c|}{30.41}          & 35.00          & \multicolumn{1}{c|}{37.24}          & 46.88          & \multicolumn{1}{c|}{51.39}          & 42.88                \\
\multicolumn{1}{l|}{MetaNER}                                      & \multicolumn{1}{c|}{770M}                     & \textbf{57.40} & \multicolumn{1}{c|}{\textbf{63.45}} & 31.59          & \multicolumn{1}{c|}{\textbf{36.52}} & \textbf{40.01} & \multicolumn{1}{c|}{\textbf{44.92}} & \textbf{52.07} & \multicolumn{1}{c|}{54.87}          & \textbf{47.60}       \\ \bottomrule
\end{tabular}
}
\caption{Micro-F1 scores of 1-shot and 5-shot in-context NER on test set. For a fair comparison, the results of each model are based on a single frozen model without fine-tuning and the pre-trained NER models are pre-trained using the same dataset as MetaNER.}

\label{tab:mainresult}
\end{table*}

To learn the above extraction ability, we design two extraction pre-training tasks, including an entity extraction task and a pseudo extraction language modeling task:

\paragraph{Entity Extraction Task.} This task is used to train the ability to extract entities from text, we use both in-context NER settings whose input is (instruction, demonstrations, text) and traditional NER settings whose input is (instruction, text), and output is entities. Note that type anonymization is only conducted in in-context NER setting.

\paragraph{Pseudo Extraction Language Modeling Task}. Because there is a mismatch between the entity extraction task and the original language modeling task, and the size of the NER corpus is usually far smaller than the text corpus for language modeling pre-training, we design a pseudo extraction LM task to bridge the above gap. Specifically, we randomly sample unlabeled sentences from the text corpus and automatically build pseudo extraction (instruction, demonstrations, text, pseudo entities) tasks. For instance, given a demonstration sentence such as ``\textit{I think this movie is cool and I really like it very much}'' and a text ``\textit{I do not like it.}'': 
(1) To begin with, we choose some spans from demonstrations (such as "this movie" and "like") and designate them as pseudo entities\footnote{We introduce how to select spans in Appendix.}. We assign random types to these entities from type indicators. For instance, we consider "this movie" as a pseudo entity of type <type2> and "like" as a pseudo entity of type <type14>.
(2) The input of the pseudo extraction task is instruction="Target types:<type2>; <type14>"; the demonstrations="Text: \textit{[MASK1] is cool and I really [MASK2] it [MASK3]}. Entities: \textit{[MASK1] is <type2>. [MASK2] is <type14>}" where the entities (``this movie'' and ``like'') and other random spans (``very much'') in demonstrations are masked. The text=``Text: I do not like it.'' which is not masked.
(3) The output of the pseudo extraction task is ``\textit{like is <type14>}'' since the model will learn from demonstrations that <type14> corresponds to "like".
(4) We also conduct traditional NER settings whose input is (instruction, text). The entities in the text will be masked as in demonstrations, e.g. ``Target types: this movie; like Text: \textit{I [MASK1] not [MASK2] it.}''. The output will be ``Entities: \textit{[MASK2] is like.}''.

We can see that the pseudo extraction LM task can benefit in-context NER in two ways. Firstly, it can significantly increase the size and diversity of in-context NER pre-training tasks from a large-scale unlabeled corpus. Secondly, this task pre-trains PLMs with a mixture of extraction target and span prediction task, therefore avoiding PLMs overfit to only extraction task. 

When pre-training, We transformed the NER and language model tasks into a uniform format and sampled input instances alternately.
\section{Experiments}
This section evaluates our method by conducting experiments on few-shot NER settings.

\subsection{Experimental Settings}
\paragraph{Pre-training settings.}
Following \citet{chen-etal-2022-shot}, we build a large-scale distant NER dataset by aligning Wikipedia and Wikidata. Specifically, our dataset was made from Wikipedia text with hyperlinks to Wikidata, where we labeled entity types using the linked Wikidata item's attributes. Entity types were gathered from Wikidata's SubclassOf and InstanceOf attributes for each span. We filtered ambiguous and low-frequency types (occurrences <100k) to obtain higher-quality demonstrations.
Finally, we retained 2046 types and 55 million (text, entities) pairs and use a 40/15 million split for training/validation.
We sample 5 million in-context tasks for training and 10k for validation, where each instance with type number $N$ is 10 and instance number $K$ is 10. 
We employ the T5-v1.1-large~\citep{raffel2020exploring} model as the initial model for MetaNER and further pre-train 500k steps with learning rate=5e-5 and warm-up steps=10k. In this paper, we refer to the pre-trained model as \textbf{MetaNER}.

\paragraph{Few-shot settings.}
Our experiments follow the standard k-shot NER setting ~\citet{huang-etal-2021-shot}: For each entity type, we sample $k$ training instances as in-context demonstrations. We evaluate models by micro-F1 and report the average performance by repeating each experiment 10 times.

We conducts experiments on 4 datasets across differnt domains: (1) CoNLL03~\citep{conll} from news domain. (2) WNUT17~\citep{wnut} from social media domain. (3) NCBI-disease~\citep{dougan2014ncbi} from biology domain. (4) SEC-filings~\citep{alvarado2015domain} from finance domain.

\paragraph{Baselines.}
For fair comparison, we use frozen models for all baselines in the in-context learning experiments, i.e., a pre-trained language/NER model is used for entity extraction without fine-tuning.
In addition, we will discuss fine-tuning based methods in section~\ref{sec:finetune}.
Two kinds of baselines are compared: 

1) \textbf{Pre-trained language models} include models with different scales and architectures: (1) Encoder-decoder models -- T5 models~\citep{raffel2020exploring}, includes T5-v1.1-large (770M), T5-xl (3B) and T5-xxl (11B). (2) Causal LM models -- GPT and OPT models~\citep{radford2019language,zhang2022opt}, includes GPT2-xl (1.5B), GPT-j-6B~\citep{gpt-j}, GPT-Neox-20B~\citep{gpt-neox-20b}, OPT-13B, OPT-30B and OPT-66B. Notice that, for PLMs, we use original type names rather than type indicators to capture the label semantics. For encoder-decoder models like T5, we formulate in-context NER as a span corruption task and the model will generate the extraction task. For example, for input ``Target entity types: disease. Text: COVID-19 is spreading. Entities: COVID-19 is disease. Text: HIV is spread by three main routes. Entities: <extra\_id\_0>'', the span corruption task requires the decoder to generate the extraction result ``<extra\_id\_0> HIV is disease.''.

2) \textbf{Pre-trained NER models} are metric-based few-shot methods, includes prototype network (ProtoNet)~\citep{snell2017prototypical}, NNshot~\citep{yang-katiyar-2020-simple}, StructShot~\citep{yang-katiyar-2020-simple} and CONTAINER~\citep{das-etal-2022-container}. We employed BERT-Large~\citep{devlin-etal-2019-bert} as the backbone and pre-trained them using the same dataset as MetaNER. For a fair comparison, we also pre-train a 220M T5-v1.1-base~\citep{raffel2020exploring} model with our meta-function pre-training algorithm (MetaNER-base).

\subsection{Main Results}
The experimental results are shown in Table~\ref{tab:mainresult}. We can see that:

1) \textbf{Few-shot NER is challenging even for large language models, while MetaNER can achieve good in-context NER performance.} Compare with best-performed PLMs, MetaNER achieves 8.4\% F1 improvements. Moreover, due to the gap between language model task and NER task, large language models achieve poor in-context learning performance on some datasets. 

2) \textbf{Our in-context NER method can achieve robust performance, even under a large source-target domain gap.} Compared with best-performed metric-based NER models, MetaNER-base and MetaNER achieves 26.8\% and 40.7\% F1 improvement.
Specifically, the performance improvement is more significant when source-target domain gap is larger, i.e., the NCBI-disease (biology domain) and SEC-filings (finance domain).

3) \textbf{Meta-function pre-training can effectively inject in-context learning ability into both small and large PLMs.}
Both MetaNER-base and MetaNER achieve impressive performance in 1-shot and 5-shot settings, which verified that MetaNER can effectively inject in-context NER ability into small PLMs, although currently in-context learning has been seen an ability only emerged only on large language models such as GPT-3.

\subsection{Detailed Analysis}
\subsubsection{Ablation Studies}
\begin{table}[th!]
\centering
\resizebox{0.45\textwidth}{!}{
\begin{tabular}{@{}l|ccc|ccc@{}}
\toprule
            & \multicolumn{3}{c|}{CoNLL03}                     & \multicolumn{3}{c}{NCBI-disease}                 \\ \cmidrule(l){2-7} 
            & P              & R              & F1             & P              & R              & F1             \\ \midrule
MetaNER     & 73.59          & 57.19          & \textbf{64.34} & \textbf{54.96} & \textbf{36.85} & \textbf{43.79} \\
w/o MF      & 68.97          & 57.62          & 62.77          & 38.27          & 35.26          & 36.28          \\
w/o LM      & 70.86          & \textbf{57.99} & 63.77          & 37.54          & 34.82          & 35.67          \\
w/o anonymization & \textbf{74.75} & 52.86          & 61.93          & 47.47          & 35.30          & 40.48          \\ \bottomrule
\end{tabular}}
\caption{Ablation studies on dev set. The results are based on 5-shot setting.}
\label{tab:abl}
\end{table}
To analyze and understand the effect of type anonymization, meta-function pre-training, entity extraction pre-training, and pseudo extraction LM pre-training, we conduct the following ablation experiments: (1) MetaNER w/o MF: remove the meta-function pre-training; (2) MetaNER w/o LM: remove pseudo extraction LM pre-training; (3) MetaNER w/o anonymization: we use the original entity type names in both pre-training and in-context NER, without using type anonymization. The results are shown in Table \ref{tab:abl}, we can see that:

1) \textbf{meta-function pre-training is critical for in-context learning ability.} By removing the meta-function pre-training, the results drop significantly when the domain gaps are larger, i.e., NCBI-disease. At the same time, meta-function pre-training is helpful for the model to make more precise predictions.

2) \textbf{The pseudo extraction LM task significantly benefits in-context NER.} We found MetaNER w/o LM results in a performance drop than MetaNER. We believe this is because, although using an automatically constructed pseudo dataset,  this task can significantly improve the size and the diversity of in-context NER tasks, meanwhile can retain a good language modeling ability. 

3) \textbf{Type name anonymization prevents in-context NER model from type name overfitting, and therefore enhances the in-context learning ability.} The ablation of type name anonymization results a 5.7\% performance drop in Table \ref{tab:abl}. We believe this is because type names will let models tend to memorize entity knowledge using type names, thus the model will not learn to capture entity knowledge from demonstrations on-the-fly.

\subsubsection{Effects of Meta-function Pre-training}
\begin{figure}[t!]
\centering 
\includegraphics[width=0.4\textwidth]{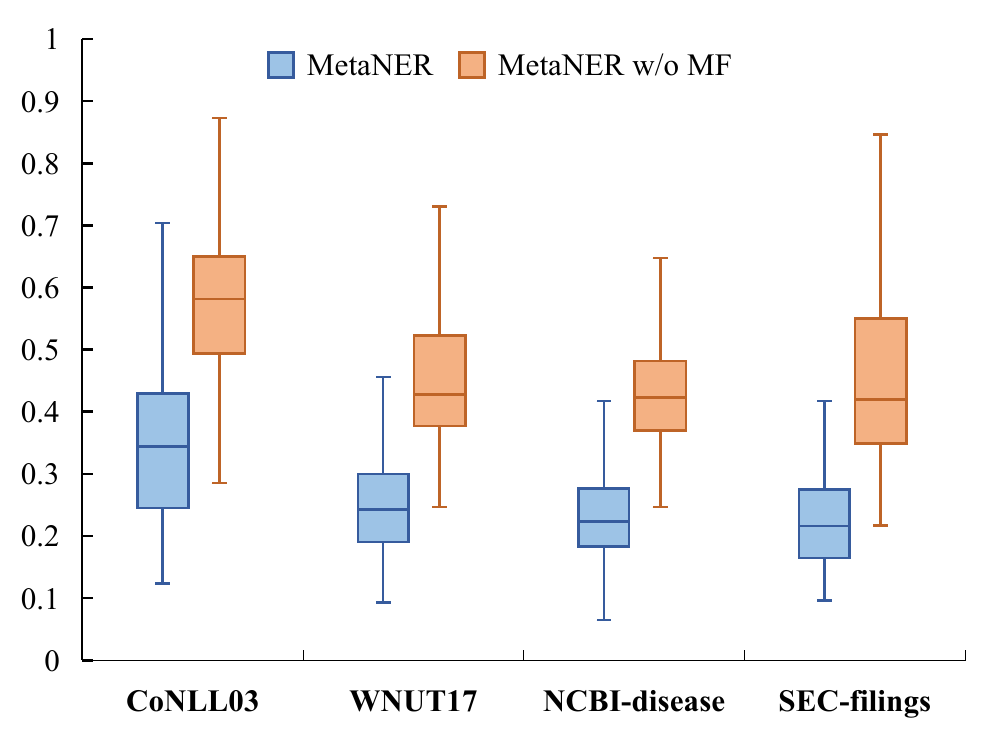}
\caption{The visualization of feature comparison between meta-function $\mathcal{F}$ and the surrogate extractor $\mathcal{F}'$. The x-axis represents the different datasets, and the y-axis represents the distances between the features from the original encoder and the features from the surrogate encoder.}
\label{Fig:vision}
\end{figure}
One main idea of this paper is that in-context NER model can be viewed as a meta-function which can implicitly build new entity extractors. To demonstrate whether meta-function pre-training can train a good meta-function, we sample 1000 instances from each dataset, and show the difference between the (instruction, demonstrations)-initialized entity extractor $\mathcal{F}$ and the surrogate entity extractor $\mathcal{F}'$, i.e.,  ||$\mathcal{F}' - \mathcal{F}$|| in Section \ref{sec:mf} in Figure \ref{Fig:vision}. We can see that meta-function pre-training can equip PLMs with a good meta-function ability, i.e., the (instruction, demonstrations)-initialized entity extractor after pre-training is significantly close to its fine-tuned counterpart.

\begin{table}[th!]
\centering
\resizebox{0.45\textwidth}{!}{
\begin{tabular}{l|cc|cc}
\toprule
                                             & \multicolumn{2}{c|}{CoNLL03}    & \multicolumn{2}{c}{WNUT17}      \\ \cmidrule(l){2-5} 
                                             & 1shot          & 5shot          & 1shot          & 5shot          \\ \midrule
BERT-large~\citep{devlin-etal-2019-bert}     & 14.66          & 52.43          & 8.95           & 32.77          \\
T5-v11-large~\citep{raffel2020exploring}     & 11.65          & 42.13          & 12.51          & 39.54          \\
GPT-NEO-20B~\citep{gpt-neox-20b}*           & 52.68          & 58.12          & 36.29          & 35.68          \\ \midrule
UIE-large~\citep{lu-etal-2022-unified}       & 46.28          & 67.62          & 32.86          & 42.67          \\
SDNet~\citep{chen-etal-2022-shot}            & /              & 71.40          & /              & 44.10          \\
CONTAINER-FT~\citep{das-etal-2022-container} & 48.56          & 66.45          & 19.46          & 24.95          \\
MetaNER-ICL*                                 & 57.40          & 63.45          & 31.59          & 36.52          \\
MetaNER-FT                                    & \textbf{61.51} & \textbf{72.70} & \textbf{39.68} & \textbf{47.26} \\ \bottomrule
\end{tabular}
}
\caption{The experiments of fine-tuning based methods. * indicates in-context learning settings. CONTAINER is pre-trained using the same NER dataset as MetaNER. All the models are implemented by us except SDNet.}
\label{tab:finetune}
\end{table}

\subsubsection{In-context Learning vs Fine-tuning} \label{sec:finetune}
MetaNER can also be directly fine-tuned using traditional NER instances. We employed the identical fine-tuning approach as previous works~\citep{huang-etal-2021-shot,lu-etal-2022-unified,chen-etal-2022-shot}. Following~\citet{lu-etal-2022-unified}, we also implemented the \textit{Rejection
Mechanism} when fine-tuning the T5-v11-large and MetaNER to achieve better few-shot performance.

To compare in-context NER with fined-tuned NER, Table~\ref{tab:finetune} reports the performance of the fine-tuned counterpart of MetaNER -- MetaNER-FT(its training is similar to surrogate entity extractor but with multi-step gradient descent until coverage), together with several fine-tuned few-shot NER baselines. We can see that: 1) MetaNER is an effective architecture, which achieves good performance on both in-context learning and fine-tuning settings; 2) Currently, fine-tuning can achieve better performance than their in-context learning counterpart. We believe this is because fine-tuned models' parameters need to be specialized to specific entity types, meanwhile in-context learning needs to generalize to different types on-the-fly, i.e., generalization-specialization trade-off. We believe this also verified the reasonableness of using a fine-tuned surrogate extractor to approximate the golden extractor.

\section{Conclusion}
In this paper, we propose an in-context learning-based NER approach and model PLMs as a meta-function, which can inject in-context NER ability into PLMs and recognize entities of new types on-the-fly using only a few demonstrative instances. Experimental results show that our method is effective for in-context NER. For future work, we will extend our method to different NLP tasks like event extraction and relation extraction.
\section*{Limitations}
In-context learning is an useful ability, this paper only focuses on in-context named entity recognition, leaves the learning of other NLP tasks' in-context learning abilities for future work.

Currently, we learn in-context learning via meta-function pre-training, by comparing an in-context extraction function and a fined-tuned surrogate extraction function at the representation level of their encoders. There are two approximation here: one is fined-tuned surrogate extraction function for approximating golden extraction function, and the difference between representations for approximating the divergence between functions. We think the above two approximations can be further improved for better and faster in-context learning.

\section*{Acknowledgements}
We sincerely thank the reviewers for their insightful comments and valuable suggestions. This research work is supported by the CAS Project for Young Scientists in Basic Research under Grant No.YSBR-040 and the National Natural Science Foundation of China under Grants no. 62122077, 62106251.

\bibliography{acl2023}
\bibliographystyle{acl_natbib}

\appendix

\section{Experiment Details}
\label{sec:appendix}

\subsection{Datasets for the extraction language model task}

Rather than randomly generating spans to form target labels in instruction, we use informative spans~\citep{bian2021bridging} as target labels. Unlike informative span selection at passage level for MRC ~\cite{bian2021bridging}, we select informative spans at a cross-document level. Specifically, we take 10 Wikipedia documents as a set and select informative spans according to the following rules: (1) spans that have appeared simultaneously in at least two and at most five documents. (2) spans that have appeared in only one document but have appeared in more than two. Rule (1) avoids some low-information general spans, such as stop words, and rule (2) retains some important spans in each document. Note that we consider at most 4-gram as a span and select the target labels from the informative spans during pre-training.

\subsection{Cost of pre-training}

We used one A-100 80g GPU for pre-training the base/large model, which took approximately one to three days. The total FLOPs for the base model are 2.30e+18 and for the large model are 7.64e+18.

\end{document}